\pdfoutput=1

\documentclass[11pt]{article}

\usepackage[]{acl}

\usepackage{times}
\usepackage{latexsym}
\usepackage{microtype}
\usepackage{booktabs}
\usepackage{multirow}
\usepackage{caption}
\usepackage{subcaption}
\usepackage{graphicx}
\usepackage{float}
\usepackage{url}
\usepackage{tipa}

\usepackage[T1]{fontenc}

\usepackage[utf8]{inputenc}

\usepackage{microtype}

\title{Quantifying Language Variation Acoustically with Few Resources}

\author{Martijn Bartelds \\
  University of Groningen \\
  The Netherlands \\
  \texttt{m.bartelds@rug.nl} \\\And
  Martijn Wieling \\
  University of Groningen \\
  The Netherlands \\
  \texttt{m.b.wieling@rug.nl} \\}

\begin{document}
\maketitle
\begin{abstract}
Deep acoustic models represent linguistic information based on massive amounts of data.
Unfortunately, for regional languages and dialects such resources are mostly not available. However, deep acoustic models might have learned linguistic information that transfers to low-resource languages. In this study, we evaluate whether this is the case through the task of distinguishing low-resource (Dutch) regional varieties. By extracting embeddings from the hidden layers of various \texttt{wav2vec~2.0} models (including new models which are pre-trained and/or fine-tuned on Dutch) and using dynamic time warping, we compute pairwise pronunciation differences averaged over 10 words for over 100 individual dialects from four (regional) languages. We then cluster the resulting difference matrix in four groups and compare these to a gold standard, and a partitioning on the basis of comparing phonetic transcriptions. Our results show that acoustic models outperform the (traditional) transcription-based approach without requiring phonetic transcriptions, with the best performance achieved by the multilingual \texttt{XLSR-53} model fine-tuned on Dutch. On the basis of only six seconds of speech, the resulting clustering closely matches the gold standard.
\end{abstract}

\section{Introduction}

Deep acoustic models have improved automatic speech recognition (ASR) substantially in recent years \cite{schneider2019wav2vec, baevski2019vq, baevski2020wav2vec, conneau2020unsupervised}.
These models represent linguistic information based on massive amounts of data.
While these models are generally evaluated on ASR benchmarks, few studies have addressed what kind of linguistic information is represented by them.
The work of \citet{livescuprobing2021} examined information represented by the \texttt{wav2vec~2.0} model \citep{baevski2020wav2vec} across the various Transformer layers.
They showed that different layers encode different types of linguistic information.
Specifically, the initial layers appeared to be most similar to the input speech features, whereas the middle layers mostly encoded contextual information.
The final layers again turned out to be similar to the input speech features. However, the representations of the final layers changed when the model was fine-tuned, likely because task-specific information was learned.
In addition, \citet{liberman2021} investigated several deep acoustic models using phonetic probing tasks, and found that representations from these models capture information useful for distinguishing English phones. Importantly, these deep acoustic models were better able to distinguish English phones than using conventional MFCC or filterbank features. Although they evaluated the transferability of deep acoustic representations across several domains, it remains unclear whether these models learned information that transfers to other languages.
This is, however, important when working on more inclusive speech technology. Especially when resources for training these models are lacking, such as for regional languages and dialects.
In this paper, we therefore investigate if hidden layers of deep acoustic models incorporate fine-grained information, which can be used to represent differences between, and in turn distinguish, regional language varieties.  

Past work on investigating language variation has often been based on computing pronunciation distances that rely on phonetically transcribed speech \citep{nerbonne1997measuring, livescu2000lexical, heeringa2004measuring}. These (edit) distances have been found to match perceptual judgements of similarity well \citep{gooskens2004perceptive, wieling2014a}.
However, transcribing speech phonetically is time-consuming and prone to errors \citep{bucholtz2007variation, novotney2010cheap}.
While automatic approaches for computing phonetic transcriptions exist (e.g., \citealt{allosaurus}), they produce lower quality phonetic transcriptions than human transcribers do.
Additionally, (discrete) phonetic transcriptions do not capture all (continuous) aspects of human speech \citep{liberman2018}.

To mitigate these shortcomings, acoustic approaches have been developed for investigating language variation \citep{Huckvale2007, Ferragne2010, Strycharczuk2020, acoustic-measure}.
However, these studies either exclusively focused on the vowels (ignoring differences in the consonants), or were negatively influenced by non-linguistic variation in the speech signal.

Recently, \citet{bartelds2021neural} found that representations from the hidden layers of pre-trained and fine-tuned \texttt{wav2vec~2.0} (large) models are suitable to represent language variation.
They showed that these representations capture linguistic information that is not represented by phonetic transcriptions, while being less sensitive to non-linguistic variation in the speech signal. Furthermore, this approach seems to provide a better match to human perceptual judgements than phonetic transcription-based approaches. 

To investigate if \texttt{wav2vec 2.0} acoustic models (including newly trained Dutch models) learn fine-grained linguistic information that can transfer to regional languages and dialects, we will assess whether or not regional languages and dialects spoken in the Netherlands can be distinguished using these models.
Our code and newly trained models are publicly available.\footnote{\url{https://github.com/Bartelds/language-variation}}

\section{Dataset}

We use Dutch dialect pronunciation recordings from the Goeman-Taeldeman-Van Reenen-Project \citep{GTRP}. Audio recordings of hundreds of words were obtained (and manually phonetically transcribed) in the 1980s and 1990s and are available for 613 dialect varieties in the Netherlands and Belgium. Unfortunately, the hour-long audio recordings were not segmented, and the metadata with the time stamps we use to extract the audio containing individual word pronunciations were only partially available. In total, therefore, we extract the acoustic recordings (judged to be of sufficient quality) for 10 words (\textit{armen}: `arms', \textit{deeg}: `dough', \textit{draden}: `wires', \textit{duiven}: `pigeons', \textit{naalden}: `needles', \textit{ogen}: `eyes', \textit{pijpen}:  `pipes', \textit{tangen}: `pliers',  \textit{volk}: `people', \textit{vuur}: `fire') pronounced in 106 locations in the Netherlands. On average, the duration of these 10 words is only 6.3 seconds for each location. Some example pronunciations are shown in Table~\ref{tab:dat}.


\begin{table}[ht!]
    \centering
    \resizebox{7.7cm}{!}{%
    \begin{tabular}{lcccc}
    \toprule
     &\textbf{Standard} & \textbf{Frisian} & \textbf{Low Saxon} & \textbf{Limburgish} \\
     & \textbf{Dutch} & (Joure) & (Eelde) & (Echt) \\
    \midrule
    Arms & \textscripta r\textschwa m\textschwa n & \textipa j\textepsilon r\textschwa m\textschwa n & \textglotstop a\textturnr ms & \ae \textinvscr \textschwa m \\
    Dough & de\textsci x & dei\c{c} & d\textepsilon ix & deix \\
    Wires & drad\textschwa n & tr\textsci dn & dr\textopeno dn & d\textscr\textbaro i \\ 
    \bottomrule
    \end{tabular}
    }
    \caption{\label{tab:dat} Phonetic transcriptions of the words `arms', `dough', and `wires' obtained from three locations where different regional languages (Frisian, Low Saxon, and Limburgish) are spoken, as well as in Standard Dutch. The names of the locations are provided between parentheses.}
\end{table}

\section{Methods}

\begin{figure}[ht!]
     \centering
      \begin{subfigure}[b]{0.23\textwidth}
         \centering
         \includegraphics[width=\textwidth]{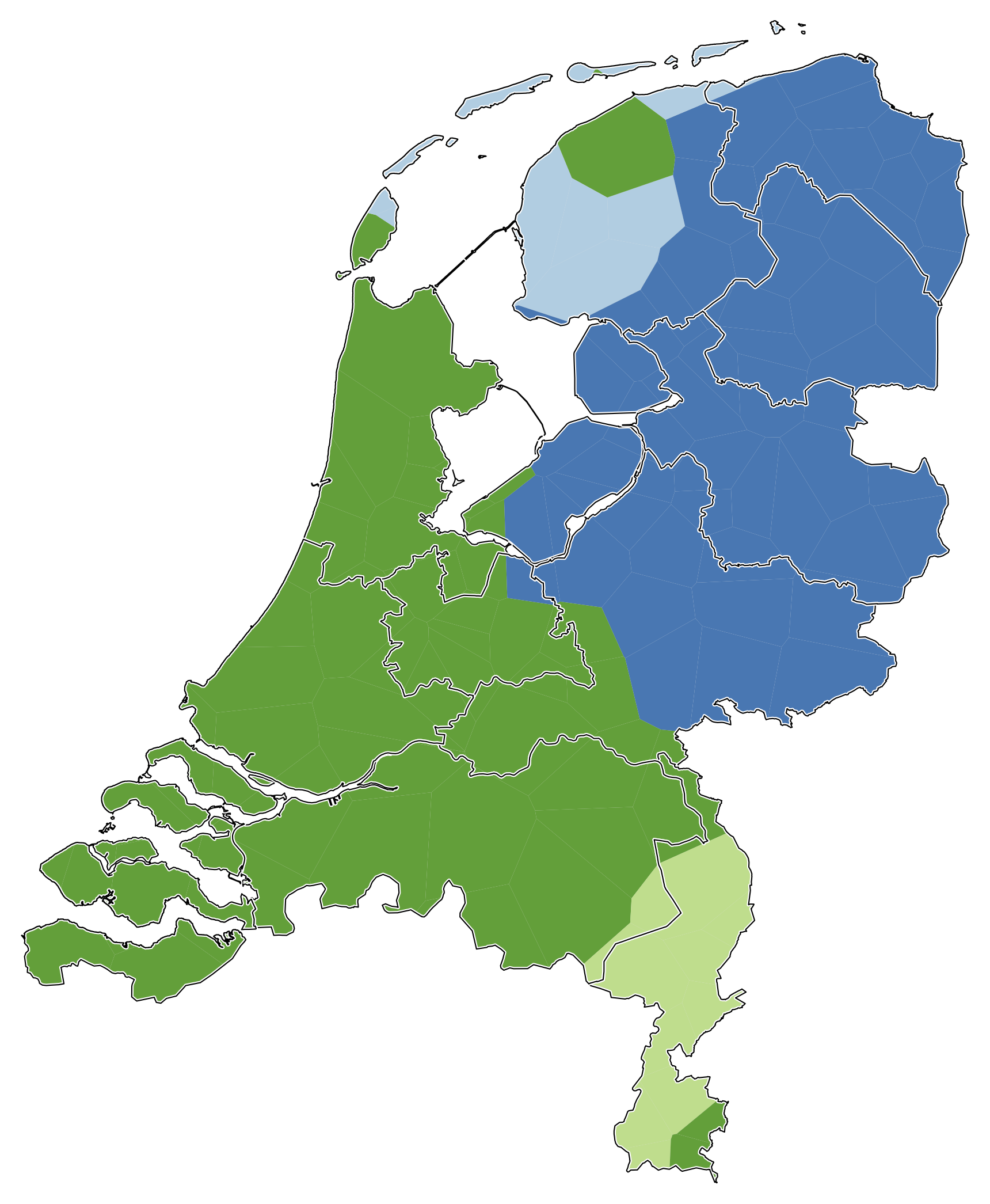}
         \caption{Gold standard clustering\\ (human-generated)}
         \label{fig:results:gold}
     \end{subfigure}
     \hfill
     \begin{subfigure}[b]{0.23\textwidth}
         \centering
         \includegraphics[width=\textwidth]{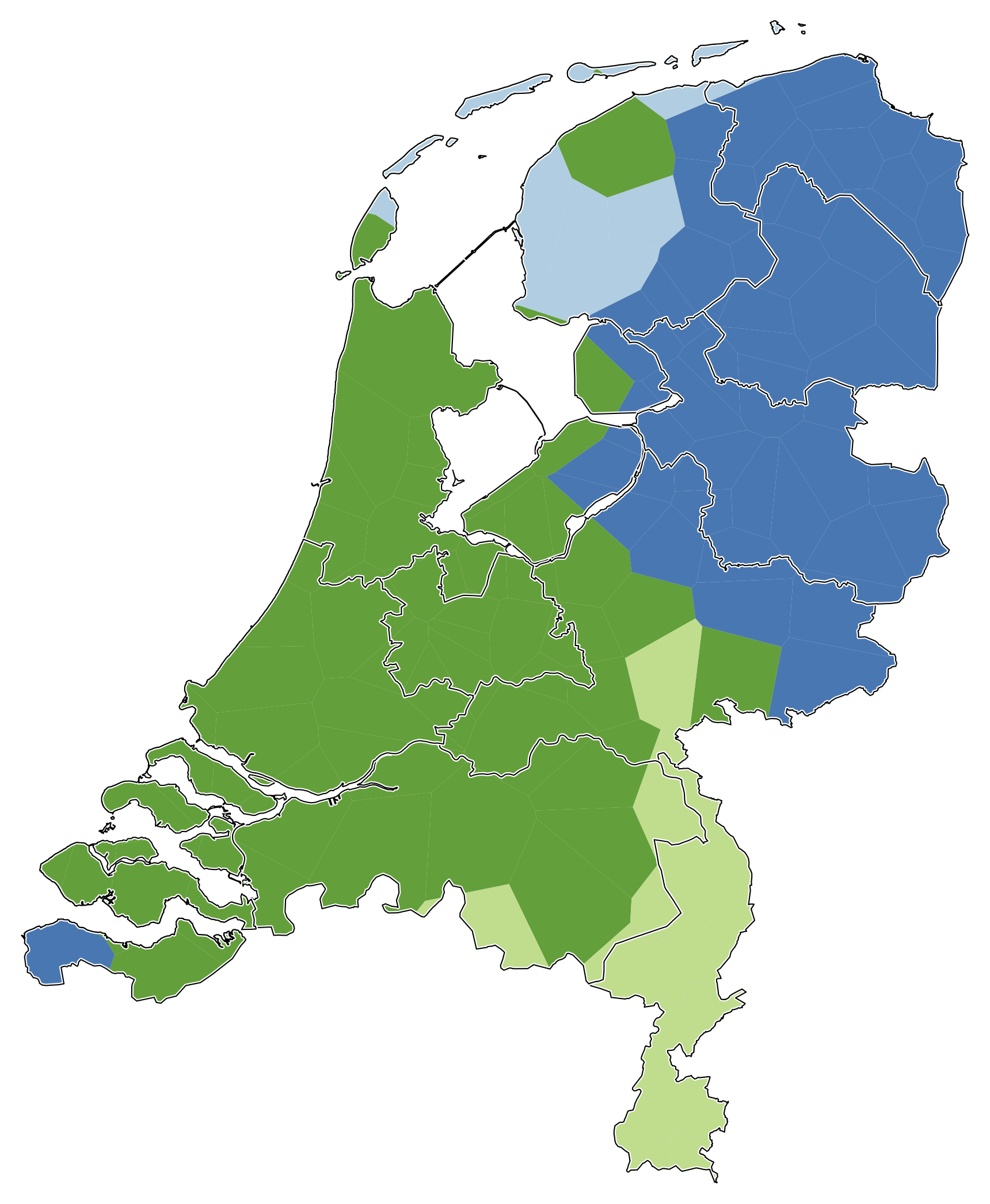}
         \caption{\texttt{XLSR-nl} layer 15\\(\texttt{cl} clustering)}
         \label{fig:results:xlsr}
     \end{subfigure}
     \hfill
      \begin{subfigure}[b]{0.23\textwidth}
         \centering
         \includegraphics[width=\textwidth]{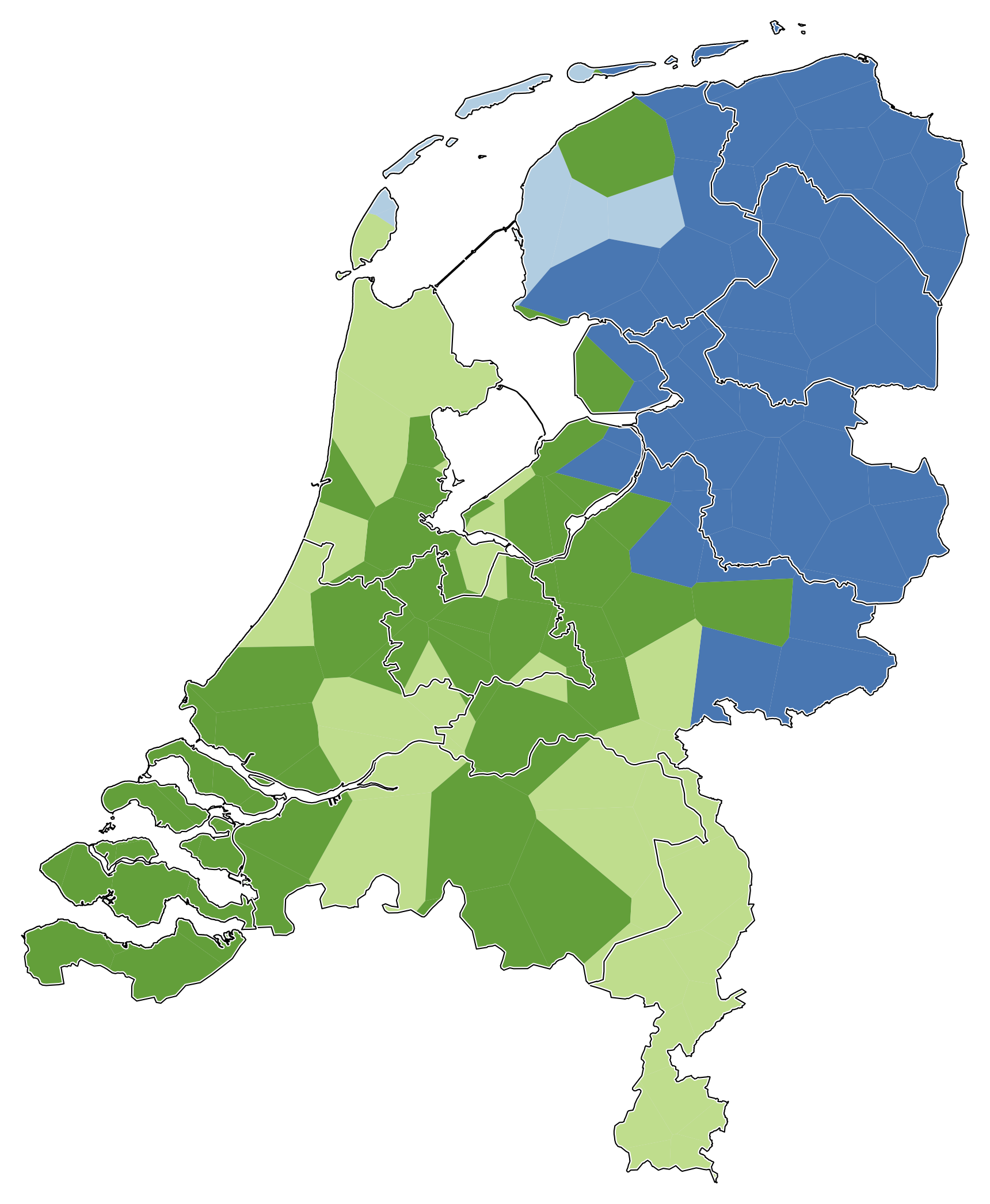}
         \caption{\texttt{w2v2-en} layer 13\\(\texttt{cl} clustering)}
         \label{fig:results:en}
     \end{subfigure}
     \begin{subfigure}[b]{0.23\textwidth}
         \centering
         \includegraphics[width=\textwidth]{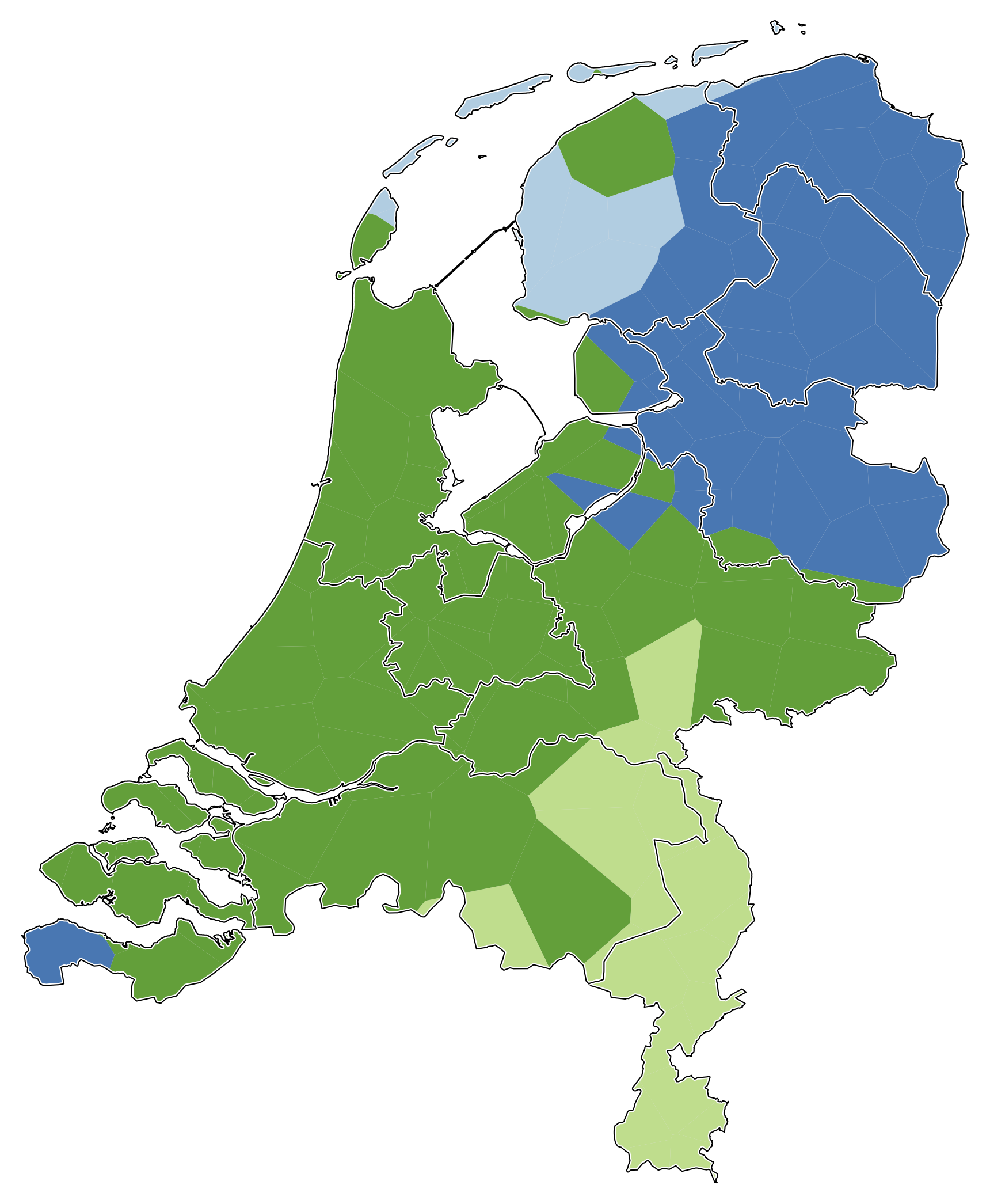}
         \caption{\texttt{w2v2-nl} layer 16\\(\texttt{wa} clustering)}
         \label{fig:results:nl}
     \end{subfigure}
     \caption{Cluster maps visualizing four clusters on the map of the Netherlands. Separate clusters are indicated by the different colours.}
     \label{fig:results:nl-xlsr-gold}
\end{figure}

We compute embeddings from the hidden Transformer layers of three fine-tuned deep acoustic \texttt{wav2vec~2.0} large models, and subsequently determine pronunciation differences using dynamic time warping (DTW) with these embeddings \citep{muller2007dynamic}.
We use fine-tuned acoustic models in this study as their hidden representations were found to show the closest match with human perceptual judgements of pronunciation variation \citep{bartelds2021neural}.
For the transcription-based approach, we apply a (phonetically sensitive) Levenshtein distance algorithm to the available corresponding phonetic transcriptions of the 10 words in all locations. After averaging the word-based differences, the result of both approaches is a distance matrix representing the aggregate pronunciation difference between every pair of locations. Both distance matrices are then clustered in four groups and quantitatively compared to a gold standard clustering of four groups (see Figure~\ref{fig:results:gold}).
These groups correspond to the three regional languages spoken in the Netherlands that are recognised by the European Charter for Regional or Minority Languages (Frisian: light blue in Figure~\ref{fig:results:gold}, Low Saxon: dark blue, Limburgish: light green) and standard Dutch (dark green).

We use the fine-tuned English \texttt{wav2vec~2.0} large model (abbreviated as \texttt{w2v2-en}) released by \citet{baevski2020wav2vec}.
In addition,  we use a new pre-trained Dutch \texttt{wav2vec~2.0} large model that is fine-tuned on Dutch labelled data (abbreviated as \texttt{w2v2-nl}), and we use the multilingual \texttt{XLSR-53} model of \citet{conneau2020unsupervised} that is fine-tuned on the same Dutch labelled data (\texttt{XLSR-nl}).
We explicitly use models for Dutch because this language is closely related to the different regional languages and dialects spoken in the Netherlands (including Frisian, Low Saxon, and Limburgish; \citealp{eberhard_ethnologue_2021}).
The advantage of having a Transformer-based language model that is linguistically closest was shown by \citet{de-vries-etal-2021-adapting}, albeit for a different task (i.e.~part-of-speech tagging). It may therefore be the case that a high degree of language similarity is also beneficial for Transformer-based models that learn speech representations. 

\paragraph{Acoustic models}
\texttt{w2v2-en} is pre-trained on 960 hours of English speech from the Librispeech dataset \citep{panayotov_librispeech_2015}.
The model consists of a convolutional encoder, a quantizer, and a 24-layer Transformer network.
Subsequently, the learned representations are fine-tuned on 960 hours of labelled data by adding a randomly initialised linear projection layer on top of the Transformer network.
This projection layer is used to predict characters from the labelled data using the connectionist temporal classification loss function (CTC; \citealp{graves2006connectionist}).

\texttt{w2v2-nl} is obtained by further pre-training the English model on 243 hours (cross-talk and silences removed) of Dutch speech from the Spoken Dutch Corpus \citep{oostdijk2000spoken}.
This approach converged faster in preliminary experiments compared to a randomly initialised network.
Subsequently, the model is fine-tuned on the same 243 hours of (now labelled) Dutch speech using CTC.
Pre-training is performed for 2 million steps with 100,000 iterations for warm up, and a linearly decreasing learning rate starting at $5\mathrm{e}{-5}$.
Fine-tuning is performed on labelled data for 1 million steps, with a linearly decreasing learning rate starting at $1\mathrm{e}{-5}$.
Other configuration details are similar to those reported in \citet{baevski2020wav2vec}.

\texttt{XLSR-53} has the same architecture as the other acoustic models, except that the quantizer has learned a single set of discrete speech representations that is shared across the pre-training languages (which includes Dutch and German, but not Frisian, Low Saxon or Limburgish).
This model is pre-trained on 56,000 hours of speech in 53 languages (44,000 hours consists of English speech) obtained from the BABEL, Common Voice and Multilingual Librispeech datasets \citep{gales2014speech, ardila2019common, pratap2020mls}.
To obtain \texttt{XLSR-nl}, \texttt{XLSR-53} is fine-tuned on the same labelled data as \texttt{w2v2-nl} with the same configuration details.


\paragraph{Obtaining pronunciation differences}
We compute pronunciation differences between all 106 locations in our dataset using both phonetic transcriptions and acoustic embeddings. For determining the phonetic transcription-based distance, we use a variant of the Levenshtein distance (\texttt{LD}) algorithm proposed by \citet{wieling2012inducing}, which includes automatically determined phonetic segment distances. This algorithm matches perception well \citep{wieling2014a} and is often used for investigating dialect variation. 

Given a pair of locations, recordings of the same word are compared using \texttt{LD} (phonetic transcriptions) or DTW (acoustic embeddings), which is a frequently-used algorithm for comparing representations of acoustic sequences \citep{senin2008dynamic}.
The acoustic embeddings are obtained for each model for each of the 24 layers separately (i.e.~to determine the optimal layer).
The word-based distances between two locations are averaged to determine the single pronunciation distance between a location pair. This process is repeated for all pairs to create a symmetric distance matrix including all locations.

\paragraph{Clustering}
We classify the phonetic transcription distance matrix and the acoustic distance matrices (three models times 24 layers) from the acoustic embeddings using seven clustering techniques, yielding the four different groups. Of course, the choice of clustering technique may influence the results, but we determine the optimal clustering algorithm by selecting the one best representing the underlying difference matrix. 
We use clustering techniques that have previously been applied to distance matrices of dialect pronunciations, namely single link (\texttt{sl}), complete link (\texttt{cl}), group average (\texttt{ga}), weighted average (\texttt{wa}), unweighted centroid (\texttt{uc}), weighted centroid (\texttt{wc}) and minimum variance (\texttt{mv}) clustering \citep{heeringa2002validating, prokic2008recognising}. 

To select the best clustering algorithm, we calculate the cophenetic correlation coefficient \citep{sokal1962comparison}. This coefficient represents the (Pearson) correlation between the original distances and the clustering-based cophenetic distances (i.e.~extracted from the dendrogram underlying the clustering). Higher values indicate a better correspondence between the original data and the clustering (with a value of 1 being perfect). 
We determine the optimal clustering method for each Transformer layer (for the acoustic models) per model by selecting the one with the highest cophenetic correlation coefficient.

\begin{figure}[t!]
     \centering
     \begin{subfigure}[ht!]{0.23\textwidth}
         \centering
         \includegraphics[width=\textwidth]{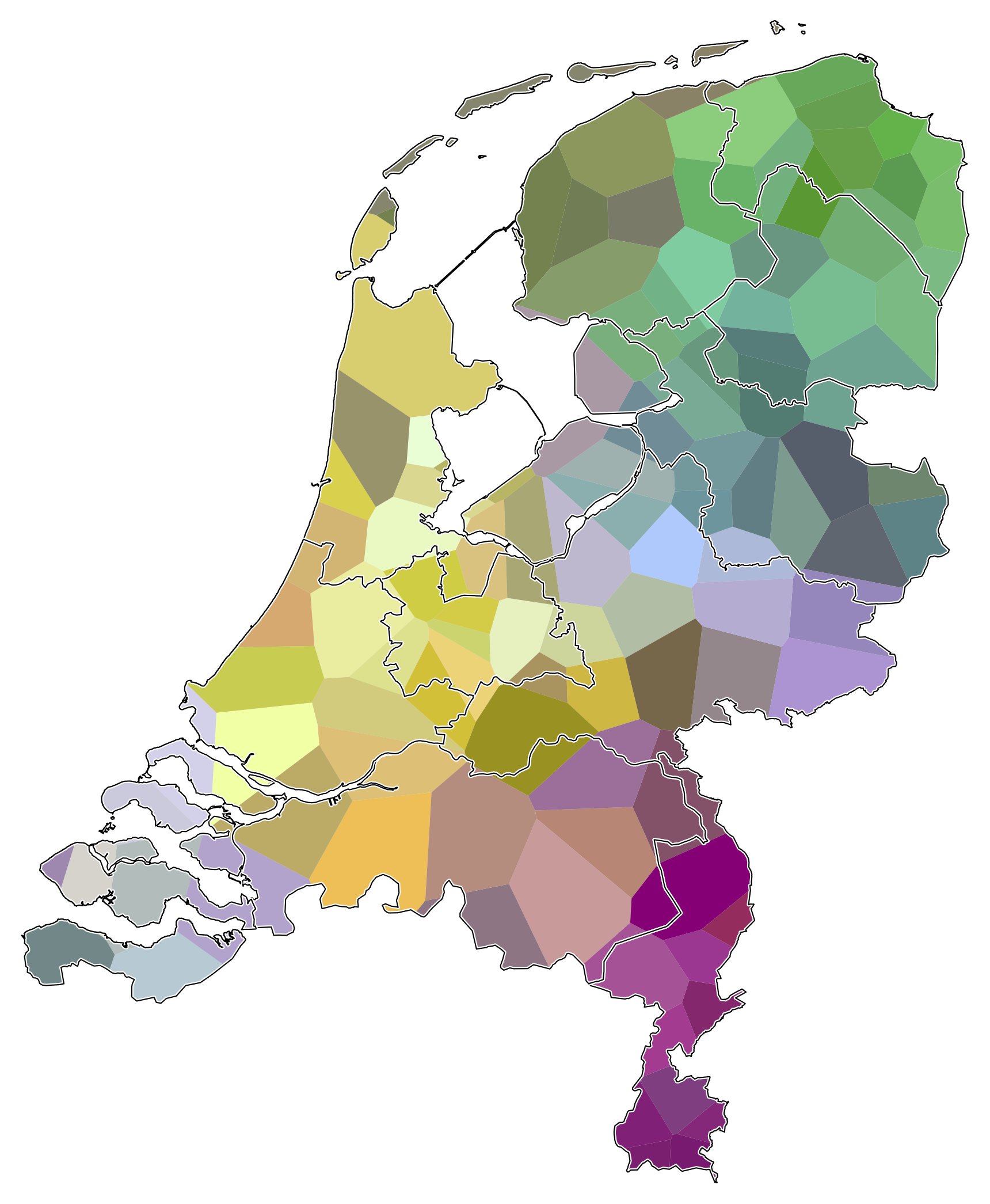}
         \caption{\texttt{XLSR-nl} layer 15}
         \label{fig:results:mds:xlsr}
     \end{subfigure}
     \hfill
     \begin{subfigure}[ht!]{0.23\textwidth}
         \centering
         \includegraphics[width=\textwidth]{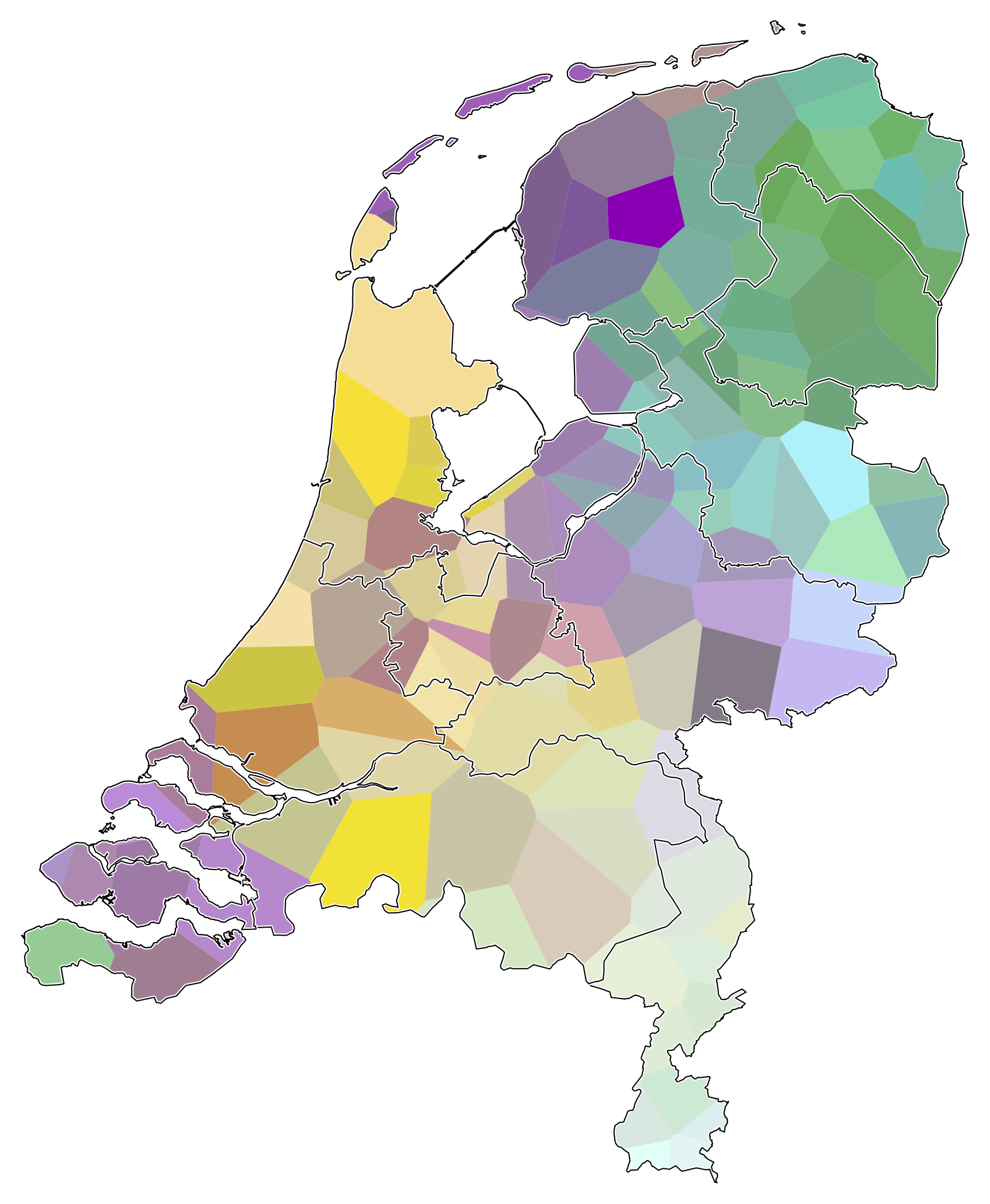}
         \caption{\texttt{LD}}
         \label{fig:results:mds:ld}
     \end{subfigure}
     \caption{MDS maps visualizing pronunciation differences based on Dutch dialect pronunciations. Similar colours correspond to pronunciations that are also similar.}
     \label{fig:results:mds:nl-xlsr-ld}
\end{figure}

\paragraph{Evaluation}
We compare the layer-based clustering results per model to the gold standard clustering.
We do this by computing the \texttt{CDistance} score, which is a clustering comparison measure proposed by \citet{coen2010comparing}.
As opposed to other techniques for comparing clustering partitions, this measure incorporates spatial information in the evaluation (i.e.~the coordinates of the locations), which is essential for evaluating spatial (i.e.~geographical) clustering. 
The \texttt{CDistance} scores (for the optimal clustering method per layer) are compared across the layers for each model. The layer with the lowest score per model (i.e.~most closely matching the gold standard clustering) is selected for the comparison of the three models. 
In addition, we create multidimensional scaling (MDS) maps \citep{torgerson1952multidimensional} using the best-performing model and compare it to the frequently used \texttt{LD} algorithm to show the (more fine-grained) relationship between the geographical location of the locations and the pronunciation differences.

\section{Results and discussion}

\begin{table}[ht!]
    \centering
    \resizebox{7.7cm}{!}{%
    \begin{tabular}{llll}
    \toprule
    \textbf{Model} & \textbf{Layer} & \textbf{Clustering} & \textbf{CDistance} \\
    \midrule
    \texttt{w2v2-en} & 13 & \texttt{cl} & 0.34 \\
    \texttt{w2v2-nl} & 16 & \texttt{wa} & 0.34 \\
    \texttt{XLSR-nl} & 15 & \texttt{cl} & \textbf{0.20} \\
    \midrule
    \texttt{LD} &  & \texttt{ga} & 0.46 \\
    \bottomrule
    \end{tabular}
    }
    \caption{\label{res:cdistance} \texttt{CDistance} scores for the different models with the optimal clustering algorithm and output layer (if applicable). Lower scores indicate a better match with the gold standard clustering.}
\end{table}

In Table~\ref{res:cdistance}, we show the \texttt{CDistance} scores associated with the different models. Ideally, the best layer would have been selected using a validation set instead of using all data, but our set of words was unfortunately too small to be adequately split. However, given that the optimal layers reported in Table~\ref{res:cdistance} correspond with the middle hidden layers found to be best representing pronunciation differences in the work of \citet{bartelds2021neural}, we do not believe this to be problematic. 

Our results show that the \texttt{XLSR-nl} model with output layer 15 and complete link clustering shows the best performance among the fine-tuned models. Note that the standard deviation of the performance for the \texttt{XLSR-nl} model across all Transformer layers was equal to 0.09, which highlights the strong performance of this model over the other models.
Importantly, all fine-tuned acoustic models improve over the \texttt{LD} algorithm, which is traditionally used to investigate (dialectal) language variation.
Perhaps surprisingly, the \texttt{w2v2-nl} model performs similar to the \texttt{w2v2-en} model. We do not have a clear explanation for this pattern, but it may be caused by the Dutch model being based on the English model, in combination with a smaller amount of Dutch as opposed to English data used for pre-training. In future work we aim to investigate this.

The multilingual \texttt{XLSR-nl} model outperforms both monolingual models. The \texttt{XLSR-nl} model is pre-trained on a variety of languages, including Dutch, English and German. The regional languages and dialects spoken in the Netherlands have clear links to these three languages (i.e.~Frisian has some overlap with English, Low Saxon has some overlap with German, and all varieties overlap with Dutch, which is also the fine-tuning language).

To illustrate, Figure~\ref{fig:results:nl-xlsr-gold} visualizes the gold standard together with the fine-tuned acoustic models. 
The \texttt{XLSR-nl} model clearly classifies pronunciations in the geographical area where Limburgish is spoken (i.e.~the light green area) most accurately. While the \texttt{XLSR-nl} model does not perfectly distinguish the Low Saxon pronunciations (i.e.~the dark blue area), the other models perform worse in this regard.

To evaluate (albeit subjectively) how well more fine-grained differences are captured by the best-performing model, Figure~\ref{fig:results:mds:nl-xlsr-ld} shows the MDS maps for the \texttt{XLSR-nl} model, as well as the \texttt{LD} algorithm.
Both approaches show the relative gradual nature of dialect variation well.
However, the \texttt{XLSR-nl} model seems to capture the larger distinctions (e.g., delineating the Limburgish area) better than the \texttt{LD} algorithm.
Based on these evaluations, \texttt{XLSR-nl} appears to be the best model when little data is available.

\section{Conclusion}
We have found that the \texttt{XLSR-nl} model can be effectively used to distinguish between language groups in the Netherlands when only a small amount of data is available. It even outperformed the \texttt{LD} algorithm, which requires time-consuming phonetic transcriptions. Our study further shows that multilingual pre-training and fine-tuning on a similar language (compared to the target languages) is beneficial over using a monolingual model.

\section*{Acknowledgements}

We gratefully acknowledge the financial support of the Center for Groningen Language and Culture (CGTC). In addition, we thank Wietse de Vries for his help in training the models, and the anonymous reviewers for their insightful feedback. Any mistakes remain our own.

\bibliography{references}

\begin{thebibliography}{37}
\expandafter\ifx\csname natexlab\endcsname\relax\def\natexlab#1{#1}\fi

\bibitem[{Ardila et~al.(2020)Ardila, Branson, Davis, Kohler, Meyer, Henretty,
  Morais, Saunders, Tyers, and Weber}]{ardila2019common}
Rosana Ardila, Megan Branson, Kelly Davis, Michael Kohler, Josh Meyer, Michael
  Henretty, Reuben Morais, Lindsay Saunders, Francis Tyers, and Gregor Weber.
  2020.
\newblock \href {https://aclanthology.org/2020.lrec-1.520} {{Common Voice: A
  Massively-Multilingual Speech Corpus}}.
\newblock In \emph{Proceedings of the 12th Language Resources and Evaluation
  Conference}, pages 4218--4222, Marseille, France. European Language Resources
  Association.

\bibitem[{Baevski et~al.(2020{\natexlab{a}})Baevski, Schneider, and
  Auli}]{baevski2019vq}
Alexei Baevski, Steffen Schneider, and Michael Auli. 2020{\natexlab{a}}.
\newblock \href {https://openreview.net/forum?id=rylwJxrYDS} {{vq-wav2vec:
  Self-Supervised Learning of Discrete Speech Representations}}.
\newblock In \emph{8th International Conference on Learning Representations,
  {ICLR} 2020, Addis Ababa, Ethiopia, April 26-30, 2020}. OpenReview.net.

\bibitem[{Baevski et~al.(2020{\natexlab{b}})Baevski, Zhou, Mohamed, and
  Auli}]{baevski2020wav2vec}
Alexei Baevski, Yuhao Zhou, Abdelrahman Mohamed, and Michael Auli.
  2020{\natexlab{b}}.
\newblock \href
  {https://proceedings.neurips.cc/paper/2020/hash/92d1e1eb1cd6f9fba3227870bb6d7f07-Abstract.html}
  {{wav2vec 2.0: {A} Framework for Self-Supervised Learning of Speech
  Representations}}.
\newblock In \emph{Advances in Neural Information Processing Systems 33: Annual
  Conference on Neural Information Processing Systems 2020, NeurIPS 2020,
  December 6-12, 2020, virtual}.

\bibitem[{Bartelds et~al.(2022)Bartelds, {de Vries}, Sanal, Richter, Liberman,
  and Wieling}]{bartelds2021neural}
Martijn Bartelds, Wietse {de Vries}, Faraz Sanal, Caitlin Richter, Mark
  Liberman, and Martijn Wieling. 2022.
\newblock \href {https://doi.org/https://doi.org/10.1016/j.wocn.2022.101137}
  {Neural representations for modeling variation in speech}.
\newblock \emph{Journal of Phonetics}, 92:101137.

\bibitem[{Bartelds et~al.(2020)Bartelds, Richter, Liberman, and
  Wieling}]{acoustic-measure}
Martijn Bartelds, Caitlin Richter, Mark Liberman, and Martijn Wieling. 2020.
\newblock \href {https://doi.org/10.3389/frai.2020.00039} {{A New
  Acoustic-Based Pronunciation Distance Measure}}.
\newblock \emph{Frontiers in Artificial Intelligence}, 3:39.

\bibitem[{Bucholtz(2007)}]{bucholtz2007variation}
Mary Bucholtz. 2007.
\newblock \href {http://www.jstor.org/stable/24049459} {Variation in
  transcription}.
\newblock \emph{Discourse Studies}, 9(6):784--808.

\bibitem[{Coen et~al.(2010)Coen, Ansari, and Fillmore}]{coen2010comparing}
Michael~H. Coen, M.~Hidayath Ansari, and Nathanael Fillmore. 2010.
\newblock \href {https://icml.cc/Conferences/2010/papers/642.pdf} {{Comparing
  Clusterings in Space}}.
\newblock In \emph{Proceedings of the 27th International Conference on Machine
  Learning (ICML-10), June 21-24, 2010, Haifa, Israel}, pages 231--238.
  Omnipress.

\bibitem[{Conneau et~al.(2020)Conneau, Baevski, Collobert, Mohamed, and
  Auli}]{conneau2020unsupervised}
Alexis Conneau, Alexei Baevski, Ronan Collobert, Abdelrahman Mohamed, and
  Michael Auli. 2020.
\newblock \href {http://arxiv.org/abs/2006.13979} {{Unsupervised Cross-lingual
  Representation Learning for Speech Recognition}}.
\newblock \emph{arXiv preprint arXiv:2006.13979}.

\bibitem[{de~Vries et~al.(2021)de~Vries, Bartelds, Nissim, and
  Wieling}]{de-vries-etal-2021-adapting}
Wietse de~Vries, Martijn Bartelds, Malvina Nissim, and Martijn Wieling. 2021.
\newblock \href {https://doi.org/10.18653/v1/2021.findings-acl.433} {{Adapting
  Monolingual Models: Data can be Scarce when Language Similarity is High}}.
\newblock In \emph{Findings of the Association for Computational Linguistics:
  ACL-IJCNLP 2021}, pages 4901--4907, Online. Association for Computational
  Linguistics.

\bibitem[{Eberhard et~al.(2021)Eberhard, Simons, and
  Fennig}]{eberhard_ethnologue_2021}
David~M. Eberhard, Gary~F. Simons, and Charles~D. Fennig. 2021.
\newblock \href {https://www.ethnologue.com} {Ethnologue: {Languages} of the
  {World}. {Twenty}-fourth edition.}
\newblock SIL International.

\bibitem[{Ferragne and Pellegrino(2010)}]{Ferragne2010}
Emmanuel Ferragne and François Pellegrino. 2010.
\newblock \href {https://doi.org/https://doi.org/10.1016/j.wocn.2010.07.002}
  {{Vowel systems and accent similarity in the British Isles: Exploiting
  multidimensional acoustic distances in phonetics}}.
\newblock \emph{Journal of Phonetics}, 38(4):526--539.

\bibitem[{Gales et~al.(2014)Gales, Knill, Ragni, and Rath}]{gales2014speech}
Mark J.~F. Gales, Kate~M. Knill, Anton Ragni, and Shakti~P. Rath. 2014.
\newblock \href {https://eprints.whiterose.ac.uk/152840/} {{Speech recognition
  and keyword spotting for low-resource languages: BABEL project research at
  CUED}}.
\newblock In \emph{Fourth International workshop on spoken language
  technologies for under-resourced languages (SLTU-2014)}, pages 16--23.
  International Speech Communication Association (ISCA).

\bibitem[{Goeman and Taeldeman({1996})}]{GTRP}
Ton Goeman and Johan Taeldeman. {1996}.
\newblock \href {http://hdl.handle.net/1854/LU-258419} {{Fonologie en
  morfologie van de Nederlandse dialecten: een nieuwe materiaalverzameling en
  twee nieuwe atlasprojecten}}.
\newblock \emph{{Taal en Tongval}}, {48}:{38--59}.

\bibitem[{Gooskens and Heeringa(2004)}]{gooskens2004perceptive}
Charlotte Gooskens and Wilbert Heeringa. 2004.
\newblock \href {https://doi.org/10.1017/S0954394504163023} {{Perceptive
  evaluation of Levenshtein dialect distance measurements using Norwegian
  dialect data}}.
\newblock \emph{Language Variation and Change}, 16(3):189–207.

\bibitem[{Graves et~al.(2006)Graves, Fern{\'{a}}ndez, Gomez, and
  Schmidhuber}]{graves2006connectionist}
Alex Graves, Santiago Fern{\'{a}}ndez, Faustino~J. Gomez, and J{\"{u}}rgen
  Schmidhuber. 2006.
\newblock \href {https://doi.org/10.1145/1143844.1143891} {Connectionist
  temporal classification: labelling unsegmented sequence data with recurrent
  neural networks}.
\newblock In \emph{Machine Learning, Proceedings of the Twenty-Third
  International Conference {(ICML} 2006), Pittsburgh, Pennsylvania, USA, June
  25-29, 2006}, volume 148 of \emph{{ACM} International Conference Proceeding
  Series}, pages 369--376. {ACM}.

\bibitem[{Heeringa(2004)}]{heeringa2004measuring}
{Wilbert} Heeringa. 2004.
\newblock \href
  {https://research.rug.nl/en/publications/measuring-dialect-pronunciation-differences-using-levenshtein-dis}
  {\emph{{Measuring Dialect Pronunciation Differences using Levenshtein
  Distance}}}.
\newblock Ph.D. thesis, University of Groningen.

\bibitem[{Heeringa et~al.(2002)Heeringa, Nerbonne, and
  Kleiweg}]{heeringa2002validating}
Wilbert Heeringa, John Nerbonne, and Peter Kleiweg. 2002.
\newblock \href
  {https://link.springer.com/chapter/10.1007/978-3-642-55991-4_48} {{Validating
  Dialect Comparison Methods}}.
\newblock In \emph{Classification, Automation, and New Media}, pages 445--452,
  Berlin, Heidelberg. Springer Berlin Heidelberg.

\bibitem[{Huckvale(2007)}]{Huckvale2007}
Mark Huckvale. 2007.
\newblock \href {https://doi.org/10.1007/978-3-540-74122-0_20} {\emph{ACCDIST:
  An Accent Similarity Metric for Accent Recognition and Diagnosis}}, pages
  258--275. Springer Berlin Heidelberg, Berlin, Heidelberg.

\bibitem[{Li et~al.(2020)Li, Dalmia, Li, Lee, Littell, Yao, Anastasopoulos,
  Mortensen, Neubig, Black, and Metze}]{allosaurus}
Xinjian Li, Siddharth Dalmia, Juncheng Li, Matthew Lee, Patrick Littell, Jiali
  Yao, Antonios Anastasopoulos, David~R. Mortensen, Graham Neubig, Alan~W.
  Black, and Florian Metze. 2020.
\newblock \href {https://doi.org/10.1109/ICASSP40776.2020.9054362} {{Universal
  Phone Recognition with a Multilingual Allophone System}}.
\newblock In \emph{2020 {IEEE} International Conference on Acoustics, Speech
  and Signal Processing, {ICASSP} 2020, Barcelona, Spain, May 4-8, 2020}, pages
  8249--8253. {IEEE}.

\bibitem[{Liberman(2018)}]{liberman2018}
Mark Liberman. 2018.
\newblock \href {https://doi.org/10.7208/chicago/9780226562599.003.0009}
  {{Towards progress in theories of language sound structure}}.
\newblock In Diane Brentari and Jackson~L. Lee, editors, \emph{Shaping
  phonology}. University of Chicago Press.

\bibitem[{Livescu and Glass(2000)}]{livescu2000lexical}
K.~Livescu and J.~Glass. 2000.
\newblock \href {https://doi.org/10.1109/ICASSP.2000.862074} {Lexical modeling
  of non-native speech for automatic speech recognition}.
\newblock In \emph{2000 IEEE International Conference on Acoustics, Speech, and
  Signal Processing. Proceedings (Cat. No.00CH37100)}, volume~3, pages
  1683--1686.

\bibitem[{Ma et~al.(2021)Ma, Ryant, and Liberman}]{liberman2021}
Danni Ma, Neville Ryant, and Mark Liberman. 2021.
\newblock \href {https://doi.org/10.1109/ICASSP39728.2021.9414776} {{Probing
  Acoustic Representations for Phonetic Properties}}.
\newblock In \emph{ICASSP 2021 - 2021 IEEE International Conference on
  Acoustics, Speech and Signal Processing (ICASSP)}, pages 311--315.

\bibitem[{M{\"u}ller(2007)}]{muller2007dynamic}
Meinard M{\"u}ller. 2007.
\newblock \href {https://doi.org/10.1007/978-3-540-74048-3_4} {{Dynamic Time
  Warping}}.
\newblock \emph{{Information Retrieval for Music and Motion}}, pages 69--84.

\bibitem[{Nerbonne and Heeringa(1997)}]{nerbonne1997measuring}
John Nerbonne and Wilbert Heeringa. 1997.
\newblock \href {https://aclanthology.org/W97-1102} {{Measuring Dialect
  Distance Phonetically}}.
\newblock In \emph{Computational Phonology: Third Meeting of the {ACL} Special
  Interest Group in Computational Phonology}.

\bibitem[{Novotney and Callison-Burch(2010)}]{novotney2010cheap}
Scott Novotney and Chris Callison-Burch. 2010.
\newblock \href {https://aclanthology.org/N10-1024} {{Cheap, Fast and Good
  Enough: Automatic Speech Recognition with Non-Expert Transcription}}.
\newblock In \emph{Human Language Technologies: The 2010 Annual Conference of
  the North {A}merican Chapter of the Association for Computational
  Linguistics}, pages 207--215, Los Angeles, California. Association for
  Computational Linguistics.

\bibitem[{Oostdijk(2000)}]{oostdijk2000spoken}
Nelleke Oostdijk. 2000.
\newblock \href {http://www.lrec-conf.org/proceedings/lrec2000/pdf/110.pdf}
  {{The Spoken {D}utch Corpus. Overview and First Evaluation}}.
\newblock In \emph{Proceedings of the Second International Conference on
  Language Resources and Evaluation ({LREC}{'}00)}, Athens, Greece. European
  Language Resources Association (ELRA).

\bibitem[{Panayotov et~al.(2015)Panayotov, Chen, Povey, and
  Khudanpur}]{panayotov_librispeech_2015}
Vassil Panayotov, Guoguo Chen, Daniel Povey, and Sanjeev Khudanpur. 2015.
\newblock \href {https://doi.org/10.1109/ICASSP.2015.7178964} {{Librispeech: An
  {ASR} corpus based on public domain audio books}}.
\newblock In \emph{2015 {IEEE} International Conference on Acoustics, Speech
  and Signal Processing, {ICASSP} 2015, South Brisbane, Queensland, Australia,
  April 19-24, 2015}, pages 5206--5210. {IEEE}.

\bibitem[{Pasad et~al.(2021)Pasad, Chou, and Livescu}]{livescuprobing2021}
Ankita Pasad, Ju-Chieh Chou, and Karen Livescu. 2021.
\newblock \href {https://doi.org/10.1109/ASRU51503.2021.9688093} {{Layer-Wise
  Analysis of a Self-Supervised Speech Representation Model}}.
\newblock In \emph{2021 IEEE Automatic Speech Recognition and Understanding
  Workshop (ASRU)}, pages 914--921.

\bibitem[{Pratap et~al.(2020)Pratap, Xu, Sriram, Synnaeve, and
  Collobert}]{pratap2020mls}
Vineel Pratap, Qiantong Xu, Anuroop Sriram, Gabriel Synnaeve, and Ronan
  Collobert. 2020.
\newblock \href {https://doi.org/10.21437/Interspeech.2020-2826} {{MLS: A
  Large-Scale Multilingual Dataset for Speech Research}}.
\newblock In \emph{Proc. Interspeech 2020}, pages 2757--2761.

\bibitem[{Prokić and Nerbonne(2008)}]{prokic2008recognising}
Jelena Prokić and John Nerbonne. 2008.
\newblock \href {https://doi.org/10.3366/E1753854809000366} {{Recognising
  Groups among Dialects}}.
\newblock \emph{International Journal of Humanities and Arts Computing},
  2(1-2):153--172.

\bibitem[{Schneider et~al.(2019)Schneider, Baevski, Collobert, and
  Auli}]{schneider2019wav2vec}
Steffen Schneider, Alexei Baevski, Ronan Collobert, and Michael Auli. 2019.
\newblock \href {https://doi.org/10.21437/Interspeech.2019-1873} {{wav2vec:
  Unsupervised Pre-Training for Speech Recognition}}.
\newblock In \emph{Proc. Interspeech 2019}, pages 3465--3469.

\bibitem[{Senin(2008)}]{senin2008dynamic}
Pavel Senin. 2008.
\newblock \href {https://csdl.ics.hawaii.edu/techreports/2008/08-04/08-04.pdf}
  {{Dynamic Time Warping Algorithm Review}}.
\newblock \emph{Information and Computer Science Department University of
  Hawaii at Manoa Honolulu, USA}, 855(1-23):40.

\bibitem[{Sokal and Rohlf(1962)}]{sokal1962comparison}
Robert~R. Sokal and F.~James Rohlf. 1962.
\newblock \href {http://www.jstor.org/stable/1217208} {{The Comparison of
  Dendrograms by Objective Methods}}.
\newblock \emph{Taxon}, 11(2):33--40.

\bibitem[{Strycharczuk et~al.(2020)Strycharczuk, López-Ibáñez, Brown, and
  Leemann}]{Strycharczuk2020}
Patrycja Strycharczuk, Manuel López-Ibáñez, Georgina Brown, and Adrian
  Leemann. 2020.
\newblock \href {https://doi.org/10.3389/frai.2020.00048} {{General Northern
  English. Exploring Regional Variation in the North of England With Machine
  Learning}}.
\newblock \emph{Frontiers in Artificial Intelligence}, 3:48.

\bibitem[{Torgerson(1952)}]{torgerson1952multidimensional}
Warren~S. Torgerson. 1952.
\newblock \href {https://doi.org/10.1007/BF02288916} {{Multidimensional
  scaling: I. Theory and method}}.
\newblock \emph{Psychometrika}, 17(4):401--419.

\bibitem[{Wieling et~al.(2014)Wieling, Bloem, Mignella, Timmermeister, and
  Nerbonne}]{wieling2014a}
Martijn Wieling, Jelke Bloem, Kaitlin Mignella, Mona Timmermeister, and John
  Nerbonne. 2014.
\newblock \href {https://doi.org/https://doi.org/10.1163/22105832-00402001}
  {{Measuring Foreign Accent Strength in English: Validating Levenshtein
  Distance as a Measure}}.
\newblock \emph{Language Dynamics and Change}, 4(2):253 -- 269.

\bibitem[{Wieling et~al.(2012)Wieling, Margaretha, and
  Nerbonne}]{wieling2012inducing}
Martijn Wieling, Eliza Margaretha, and John Nerbonne. 2012.
\newblock \href {https://doi.org/https://doi.org/10.1016/j.wocn.2011.12.004}
  {{Inducing a measure of phonetic similarity from pronunciation variation}}.
\newblock \emph{Journal of Phonetics}, 40(2):307--314.

\end{thebibliography}
\bibliographystyle{acl_natbib}

\end{document}